%% file: main_cvpr.tex
\documentclass[10pt,twocolumn,letterpaper]{article}

\usepackage{cvpr}
\usepackage{times}
\usepackage{epsfig}
\usepackage{graphicx}
\usepackage{amsmath}
\usepackage{amssymb}

\newcommand{\white}[1]{\textcolor{white}{\textbf{#1}}}
\newcommand{\rcite}[1]{\textcolor{red}{\cite{}}}
\usepackage{caption}
\usepackage{stackengine}
\usepackage[caption=false]{subfig}
\usepackage{comment}
\usepackage{booktabs} 
\usepackage{boldline,multirow}
\usepackage{algorithm}
\usepackage{algorithmicx}
\usepackage[noend]{algpseudocode}
\algdef{SE}[DOWHILE]{Do}{doWhile}{\algorithmicdo}[1]{\algorithmicwhile\ #1}%

\usepackage[pagebackref=true,breaklinks=true,letterpaper=true,colorlinks,bookmarks=false]{hyperref}

\cvprfinalcopy 

\ifcvprfinal\fi
\begin{document}
	\title{Greedy Structure Learning of Hierarchical Compositional Models}
	\author{Adam Kortylewski \;\; Aleksander Wieczorek \;\; Mario Wieser \;\; Clemens Blumer \\ \;\; Sonali Parbhoo \;\; Andreas Morel-Forster\;\;Volker Roth\;\;Thomas Vetter\\
		Department of Mathematics and Computer Science, 
		University of Basel}

\maketitle
	\begin{abstract}
		In this work, we consider the problem of learning a hierarchical generative model of an object from a set of images which show examples of the object in the presence of variable background clutter. Existing approaches to this problem are limited by making strong a-priori assumptions about the object's geometric structure and require segmented training data for learning. In this paper, we propose a novel framework for learning hierarchical compositional models (HCMs) which do not suffer from the mentioned limitations. We present a generalized formulation of HCMs and describe a greedy structure learning framework that consists of two phases: Bottom-up part learning and top-down model composition. Our framework integrates the foreground-background segmentation problem into the structure learning task via a background model. As a result, we can jointly optimize for the number of layers in the hierarchy, the number of parts per layer and a foreground-background segmentation based on class labels only. We show that the learned HCMs are semantically meaningful and achieve competitive results when compared to other generative object models at object classification on a standard transfer learning dataset. 		
	\end{abstract}
	\input{introduction}
	\input{related_work}
	\input{theoretical_background}
	\input{greedy}
	\input{results}
	\input{conclusion}

	\bibliographystyle{spmpsci}

	\normalsize
	\bibliography{egbib}  

\end{document}

%% file: introduction.tex
	\section{Introduction}
\begin{figure*}
	    \centering
	    \hspace{0.5cm}
	    \subfloat[\label{fig:2inp}]{\includegraphics[height=4cm]{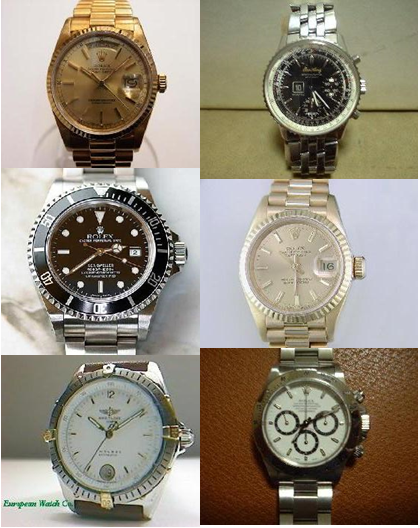}} \hspace{1.5cm} 
	    \subfloat[\label{fig:2zhu}]{\includegraphics[height=4cm]{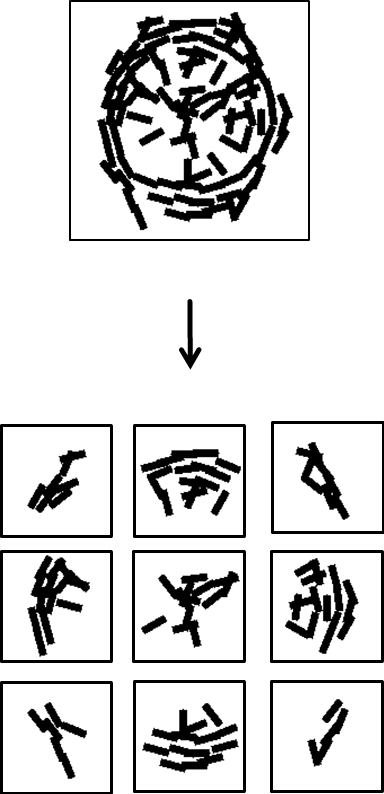}} \hspace{1.2cm} 
	    \subfloat[\label{fig:2ours}]{\includegraphics[height=4cm]{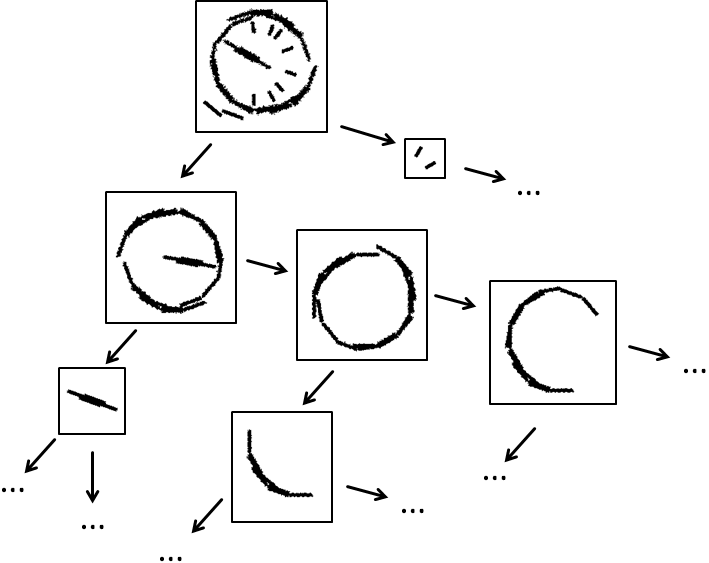}} 
		\caption{Comparison of different types of hierarchical compositional models. (a) A sample of the training data; (b \& c) Hierarchical compositional models with black strokes indicating edge features at the different location and orientation. (b) The approach as proposed by Dai et al. \cite{dai2014} learns an unnatural rather arbitrary decomposition of the object. (c) Our proposed greedy compositional clustering process learns a semantically meaningful hierarchical compositional model without the need of any a-priori knowledge about the object's geometry.}
	    \label{fig:comparisonzhu}
	    \end{figure*}

    Object analysis in natural images requires generalization from limited observations to a potentially infinite amount of image patterns that are generated by variations of an objects geometry, appearance and background clutter. Generative object modeling \cite{grenander1996elements,zhu2007stochastic,blanz1999morphable} is a highly promising approach to object analysis as it naturally integrates different analysis tasks, such as detection, segmentation and classification, into a joint reasoning process. However, so far the learning of generative object models requires detailed human supervision during training, while posterior inference at test time is slow. Hierarchical compositional generative models \cite{jin2006,fidler2014,zhu2008} proposed to resolve these issues by enforcing a more efficient representation that allows for fast inference, feature sharing and contextual reasoning. Such hierarchical compositional models (HCMs) demonstrated impressive generalization capabilities for a diverse set of applications such as image classification \cite{fidler2014}, object parsing \cite{zhu2008}, domain adaptation \cite{dai2014} and one-shot learning \cite{wong2015}. 
    However, so far HCMs can only be learned if their hierarchical structure is either known a-priori \cite{dai2014} (Figure \ref{fig:2zhu}) or if the objects in the training data are segmented from the background \cite{george2017generative, kortylewski2016probabilistic}. Therefore, a major open research question is:
	\begin{center}
	    \textit{How can the graph structure of hierarchical compositional models be learned from natural images without detailed human supervision?}
	\end{center} 
	    
	The major challenge when learning the structure of HCMs is that it requires the resolution of a fundamental chicken-and-egg problem: In order to learn the graph structure of an HCM the object must be segmented from the background, however, in order to segment the object from the background an object model is needed. Existing structure learning approaches resolve this problem by taking one of the following assumptions: 
	\begin{itemize}
	    \item[\textbf{A1}] The structure of the object is known a-priori, in terms of the number of parts and their hierarchical relation \cite{zhu2008,dai2014} (Figure \ref{fig:2zhu}).		
	    \item[\textbf{A2}] The object can be discriminated from the background solely based on local image information~\cite{fidler2014,zhu2008}.
	    \item[\textbf{A3}] The object in the training images is already segmented from the background \cite{george2017generative,kortylewski2016probabilistic}.  
	\end{itemize}
	
	The assumptions A1 and A3 are unsatisfying as they require detailed human supervision during training. Assumption A2 does not hold in general for natural images because the appearance of objects is locally highly ambiguous \cite{kersten2003bayesian}. When learning from natural images this leads to background structures becoming part of the object model, or object parts being missed in the representation \cite{spehr2014hierarchical}.
	
	The major contribution of this paper is a framework for learning the graph structure of hierarchical compositional models without relying on the assumptions A1-A3. 
	In particular, we make the following contributions:
	\begin{itemize}
		\item \textbf{Generalized formulation of HCMs.} We present a generalized formulation of hierarchical compositional models that allows for probabilistic modeling of objects with arbitrary numbers of parts.
	    \item \textbf{Greedy structure learning framework.} We propose a novel greedy learning framework for hierarchical compositional models. It consists of a bottom-up compositional clustering process that infers the number of parts per layer as well as the number of layers in a HCM. A subsequent top-down process composes the learned hierarchical parts into a holistic object model. 
		\item \textbf{Background modeling in structure learning.} We introduce a background model into the structure learning process and thus integrate the foreground-background segmentation task into the learning procedure. In this way, we can resolve the need for providing segmented training data.
	    \item \textbf{Overcoming limitations of related work.} Our qualitative results demonstrate that semantically meaningful HCMs are learned without relying on the assumptions A1-A3. Our quantitative experiments at transfer learning on the Four Domain dataset \cite{gong2012} show that our learned HCMs outperform other generative approaches in terms of classification accuracy.
	\end{itemize}

%% file: related_work.tex
	\section{Related Work}
	\label{sec:related}
	\noindent\textbf{Deformable object models:}	Deformable object models explicitly represent an object in terms of a reference object and a model of how instances of the object can deform \cite{grenander1996elements}. In their seminal work, Kaas et al. \cite{kass1988snakes} proposed an approach for detecting deformable contours in images with a hand-designed deformation model. Cootes et al. \cite{cootes1995active} were the first to learn a statistical deformation model from data. Yuille et al.  \cite{yuille1992feature} proposed to relax the global dependence between different parts of the object by introducing a hierarchical model structure. In this way, the parts could move locally independently, while, a global energy term constrained the global structure of the model. Such tree-structured models can be optimized efficiently, and therefore gained significant momentum, leading to a body of work that has developed along this line of research \cite{weber2000unsupervised,felzenszwalb2005pictorial,epshtein2005,jin2006,hinton2006,serre2007}. The Active Basis Model \cite{wu2010} is a deformable object model that is formulated within an elegant information-theoretic framework and, in addition to shape deformations, also models the object's appearance. In this work, we use the hierarchical compositional generalization of the Active Basis Model \cite{dai2014} as object representation.
        \begin{figure*}
            \centering
            \subfloat[\label{fig:cabm_graph-b}]{\includegraphics[height=4cm]{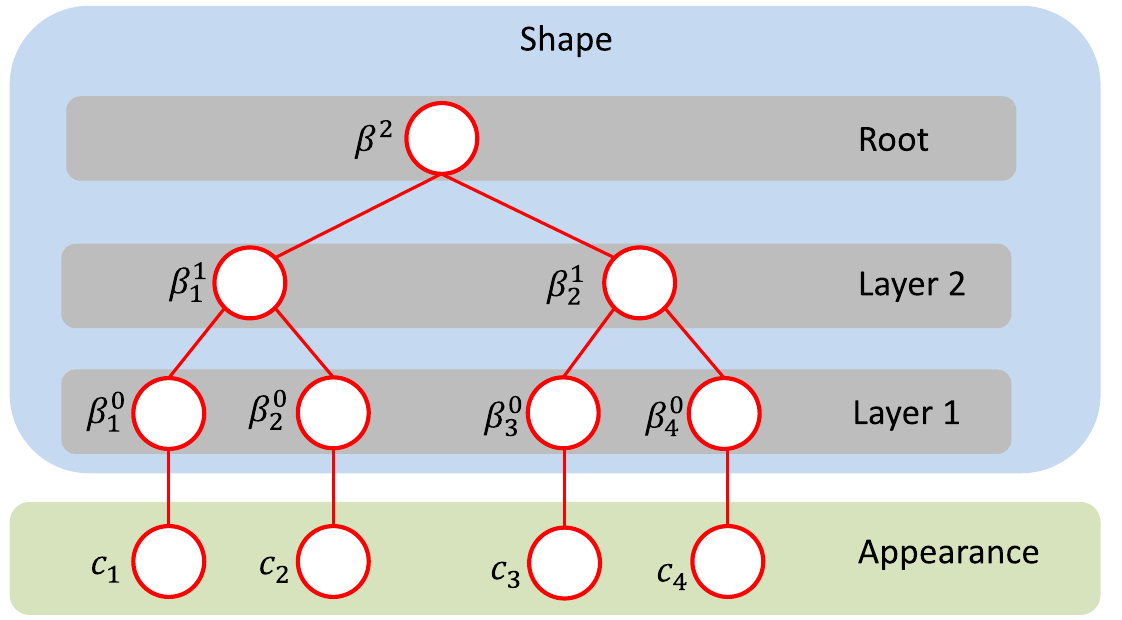}}   \hspace{1cm}         
            \subfloat[\label{fig:cabm_graph-a}]{\includegraphics[height=3.9cm]{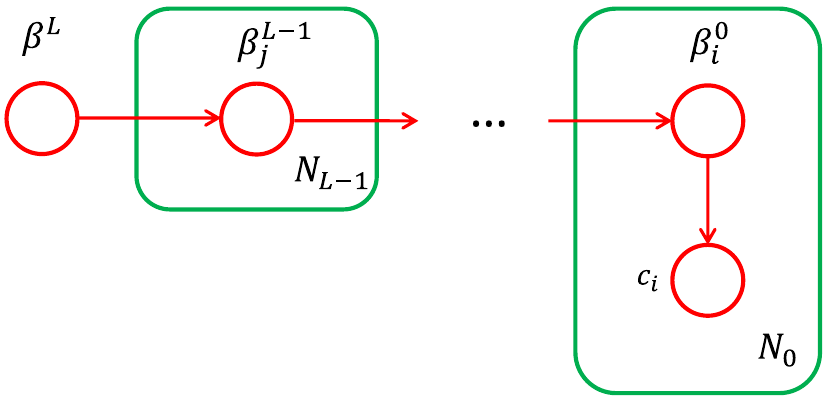}}
            \caption{The dependence structure between random variables in a Compositional Active Basis Model. (a) The simplest possible CABM, a binary-tree structured Markov random field. (b) The graphical model of a generalized multi-layer CABM (Section \ref{sec:mcabm}). We learn the full multi-layer structure of a CABM including the number of layers $L$, the number of parts per layer $N_L, \dots, N_0$ as well as their hierarchical dependence structure.}
            \label{fig:cabm_graph}
        \end{figure*}	
        
    \textbf{Hierarchical compositional models.} Hierarchical compositional models have developed as a class of models which extend deformable templates into hierarchical graphs that explicitly allow for part sharing, and thus yield big gains in computational efficiency. Furthermore, they have proven to be highly robust under strong changes in the visual domain \cite{yuille2018deep}, while achieving state-of-the-art performance in several computer vision tasks \cite{ommer2010,zhu2010,jin2006,todorovic2008,dai2014,si2013}. These approaches hand-specify the graph structures of the models and are restricted to learn the parameters only, whereas, in this work we propose to learn the graph structure from data.
    
    \textbf{Learning the structure of hierarchical compositional models.} It is desirable to learn the structure of HCMs from data. A number of works \cite{fidler2007,fidler2014,ferrari2010images,kokkinos2011inference,zhu2008} showed that exploiting the modularity of compositional models makes possible to learn HCMs one parent child clique at a time in a bottom-up manner by clustering contours. Recent work, on object recognition in changing visual environments \cite{george2017generative,kortylewski2016probabilistic,kortylewski2017model} showed that HCMs are extremely data efficient while being highly adaptive to different visual domains. These methods, however, rely on segmented training data. In contrast, we propose to learn the structure and parameters of hierarchical compositional models from natural images without relying on detailed human supervision in terms of segmented training data or prior knowledge about the object's geometry.
    
    \textbf{Background modeling for generative object models.} When analyzing images with generative object models, the background is often not modeled explicitly but represented implicitly via detection thresholds or manually specified constraints on the range of model parameters \cite{schonborn2015background}. 
    In \cite{wu2010,schonborn2015background} the authors propose to resolve such artificial constraints by letting an explicit background model compete with     the generative object model (the foreground) in explaining a target image during inference. In this paper, we propose to integrate an explicit background model as competitor of the generative object model \textit{during learning}. In this way, we integrate the foreground-background segmentation task into the learning process and thus overcome the need for detailed human supervision. 

%% file: theoretical_background.tex
	\section{Theoretical background}
	In this section, we describe the theoretical details of our hierarchical compositional object model. We start by introducing the Active Basis Model (ABM, Section \ref{sec:abm}) and its compositional generalization (CABM, Section \ref{sec:cabm}). Building on this theoretical background, we introduce the proposed generalized multi-layer CABM in Section \ref{sec:mcabm}.
	\subsection{Active Basis Model} 
	\label{sec:abm}
	ABMs \cite{wu2010}	are probabilistic generative models that model an object's variability in terms of shape and appearance. An ABM represents an image $I$ as a linear combination of basis filters $F_{\beta^0_i}$:
    \begin{equation} 
    \label{eq:3-sparse}
    I = \sum_{i=1}^{N}  c_{i} F_{\beta^0_i} + U.
    \end{equation}
    The image $I$ is decomposed into a set of Gabor filters $F_{\beta^0_i}$  with fixed frequency band, coefficients $c_i$ and a residual image $U$. The variable $\beta^0_i$ denotes the absolute position and orientation of a basis filter in the image frame. These parameters are encoded relative to the objects center $\beta^1$ such that $\beta^0_i = \Delta\beta^0_i+\beta^1$. The superscripts of the parameters indicate the layer at which the variables are located in the graphical model of the ABM. This will become important when we discuss the hierarchical generalization of ABMs in the next section. The parameters of the filters $B^0 = \{\beta^0_i | i=1,\dots,N\}$ can be learned with matching pursuit \cite{mallat1993} from a set of training images as introduced in \cite{wu2010}. By inducing a probability distribution on the model parameters $C=\{c_0,\dots,c_{N}\}$ and $B=\{B^0,\beta^1\}$ a generative object model is defined: 
    \begin{align}\label{eq:abm}
        p(C, B) & = p(\beta^1)\prod_{i=1}^N p(\beta^0_i | \beta^1) p(c_i|\beta^0_i)  .
    \end{align}
    The prior of the position and orientation of the object $p(\beta^1)$ is simply uniformly distributed over all possible rotations and all positions in the image frame. The position of the individual filters varies locally according to a uniform distribution $p(\beta_i^0 | \beta^1) = \mathcal{U}(\hat{\beta}_i^0-\delta \beta \hspace{.1cm}, \hat{\beta}_i^0 +\delta \beta )$ around the mean position of a filter $\hat{\beta}^0_i$ with $\delta \beta $ describing the possible spatial perturbation. The filter coefficients follow a statistical distribution in the form of an exponential family model: 
    \begin{equation}
    \label{eq:3-coeff}
    p(c_i|\beta^0_i)=\frac{exp(\bar{c_i} \lambda(\beta^0_i))}{Z(\lambda(\beta^0_i))},
    \end{equation}
    where each filter coefficient is bounded with a sigmoid transform $\bar{c_i}=\tau [2/(1+exp(-2c_i/\tau))-1]$ saturating at value $\tau$, in order to prevent the overfitting of the model to strong edges. The natural parameter $\lambda(\beta^0_i)$ is learned from the training data via maximum-likelihood estimation and the normalizing constant $Z(\lambda(\beta^0_i))$ can be estimated by integrating the numerator on a set of training images (more details on this process can be found in \cite{wu2010}). The core limitation of ABMs is that they assume statistical independence between individual basis filters (Eq. \ref{eq:abm}). Therefore they are limited in terms of their ability to model large object deformations and strong appearance changes \cite{dai2014}. In the next section, we introduce Compositional Active Basis Models  \cite{dai2014} which overcome this limitation by introducing hierarchical relations between the basis filters.
    \begin{figure}
       		\centering
       		\includegraphics[width=0.4\textwidth]{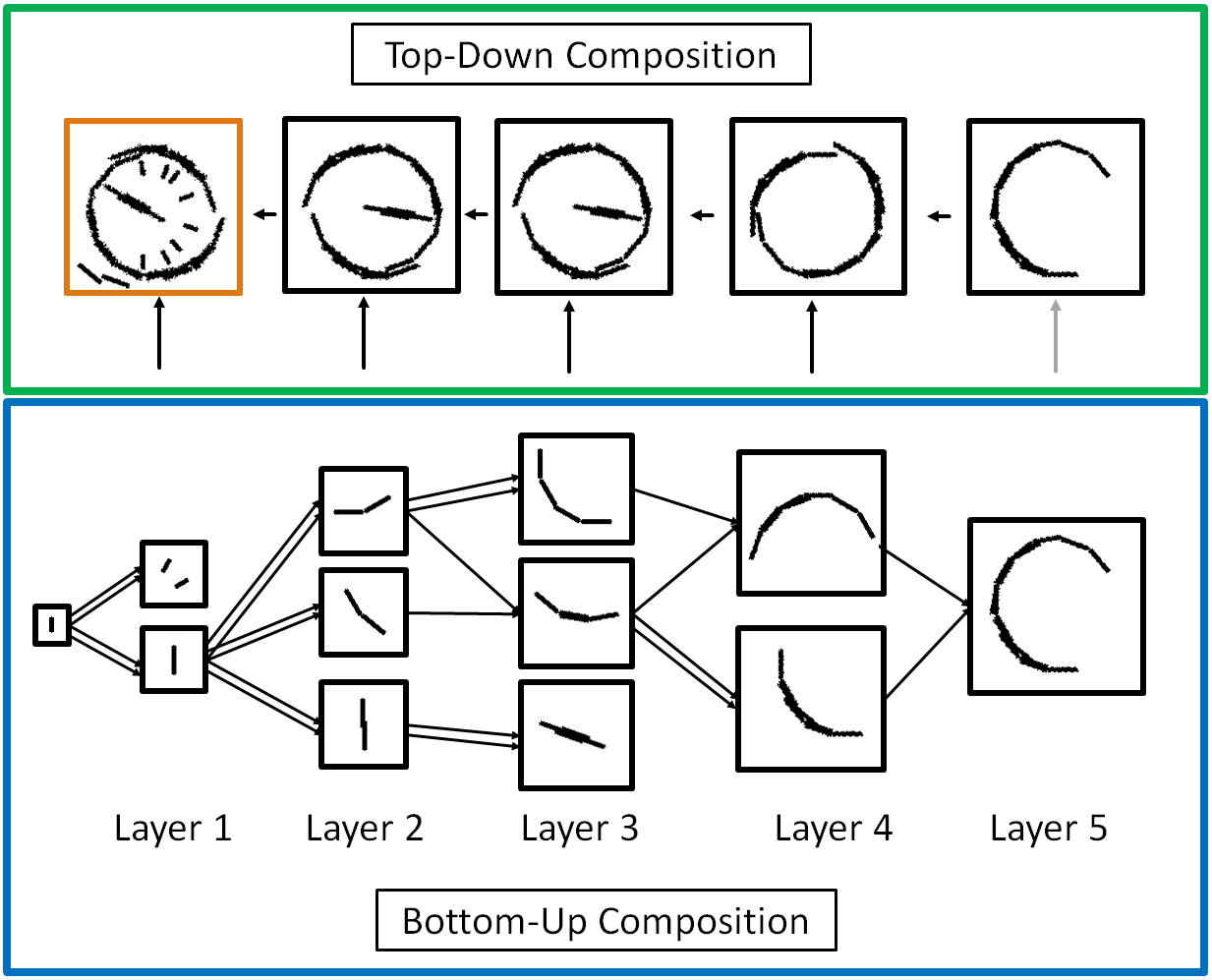}
       		\caption{Illustration of the joint bottom-up and top-down compositional learning scheme. During the bottom-up process (blue box) basis filters (black strokes) are grouped into higher-order parts until no further compositions are found. The subsequent top-down process (green box) composes the learned hierarchical part dictionary into a holistic object model (orange box).}
       		\label{fig:cabm_learn}
    \end{figure}    
  	
	\subsection{Compositional Active Basis Model} 
	\label{sec:cabm}	
	 Figure \ref{fig:cabm_graph-b} graphically illustrates the dependency structure of a two-layered Compositional Active Basis Model (CABM) as proposed by Dai et al. \cite{dai2014}. Note the tree-like dependency structure between the variables which enables a fast posterior inference via dynamic programming. The probabilistic image model of a two-layered CABM is defined as:
    \begin{equation}
    \label{eq:cabm}
    p(C, B) = p(\beta^2)\hspace{-.4cm}\prod_{j \in ch(\beta^2)} \hspace{-.3cm}p(\beta^1_j | \beta^2)\hspace{-.3cm}\prod_{i \in ch(\beta^1_j)} \hspace{-.3cm}p(\beta^0_i | \beta^1_j) p(c_i|\beta^0_i),
    \end{equation}
    where the operator $ch(\cdot)$ selects the set of children nodes. Compared to the original ABM (Equation \ref{eq:abm}), additional dependencies are introduced between group of the individual basis filters ($ch(\beta^1_j)$). In this way, the object's global structure is partitioned into multiple conditionally independent groups of basis filters. This allows for the modeling of long-range correlations in the object's geometry which cannot be achieved with the standard ABM. The learning of CABMs was originally proposed in \cite{dai2014}, however, the number of parts per layer was assumed to be known a-priori and the number of layers was fixed to 2 (see Figure \ref{fig:2zhu}). In the next section, we present a generalization of the CABM which will enable us to overcome this assumption via a greedy structure learning framework in Section \ref{sec:gcc}.
    
    \subsection{Proposed approach: Multi-Layer CABM}
   	\label{sec:mcabm}
	We can generalize the CABM model to an arbitrary numbers of hierarchical layers $L$:
    \begin{equation}
        \label{eq:mcabm}
        p(C, B) = p(\beta^L)\hspace{-.4cm}\prod_{k \in ch(\beta^L)} \hspace{-.3cm}p(\beta^{L-1}_k | \beta^L)\hspace{.1cm}\dots\hspace{-.2cm}\prod_{i \in ch(\beta^1_j)} \hspace{-.3cm}p(\beta^0_i | \beta^1_j) p(c_i|\beta^0_i),
	\end{equation}
	which corresponds to the graphical model shown in Figure~\ref{fig:cabm_graph-a}. Based on this multi-layer generalization the model becomes more expressive, and therefore can represent objects with very different geometry structure, such as e.g.\ long and thin objects as well as small but compact objects.	In this way, we evade the need for specifying the dependency structure of the model a-priori, and thus overcome the main limitation of the original model \cite{wu2010} (compare Figures~\ref{fig:2zhu}~and~\ref{fig:2ours} with a specified and learned dependence structure, respectively). However, the additional model flexibility comes at the price of having to learn the full dependency structure of the probabilistic model, including the number of layers $L$, the number of parts per layer $N_L,\dots,N_0$ and their hierarchical dependency structure. Note that Eq.~(\ref{eq:mcabm}) can be used to compute and compare posterior probabilities of models which are composed of different numbers of components. In the next section, we propose a greedy structure learning framework for estimating those parameters from data.
	
    \input{pseudo}	

%% file: pseudo.tex
\begin{algorithm}[t]
	\textbf{Input:}  Set of Gabor filters $B^0=\{\beta^0_0,\dots,\beta^0_{n_0}\}$;\\
	 \white{.}\hspace{0.85cm} Set of training images $I$\\
	\textbf{Output:} Set of Hierarchical Compositional Part Models $B=\{B^1,\dots,B^L\}$.
	\begin{algorithmic}[1]
		\State $L=1$
		\Do
			\State $B^L$ $\gets$ GreedyLearning($B^{L-1},I$)
			\State $L=L+1$			
		\doWhile{$p(B^L,\dots,B^0 | I) > p(B^{L-1},\dots,B^0 | I)$}
		\vspace{5mm}
		\Function{GreedyLearning}{$B^{L-1},I$}
		\State \hspace{-0.1cm}$n=0$
		\State \hspace{-0.1cm}$B^{L}=\{\}$
		\Do
			\State $\beta^L_n, \beta^L_{n+1} = $ init\_random\_models($B^L$)
			\State \texttt{// $\beta^L_{n+1}$ serves background model}
			\For{\#iterations}
				\State \hspace{-0.1cm}\texttt{// E-Step}
				\State \hspace{-0.1cm}\texttt{data} $\gets$ get\_training\_patches($\beta^L_0,\dots,\beta^L_{n+1},I$) 
				\State \hspace{-0.1cm}\texttt{// M-Step}
				\State \hspace{-0.1cm}$\beta^L_n$ $\gets$ learn\_compositional\_model(\texttt{data},$B^{L-1}$)
				\EndFor
			\State $B^L$ $\gets$ $\{B^L,\beta^L_n\}$
			\State n = n + 1
		\doWhile{$p(\beta^L_n,\dots,\beta^L_0 | I) > \hspace{-0.1cm}p(\beta^L_{n+1},\beta^L_{n-1},\dots,\beta^L_0 | I)$}
		\EndFunction
	\end{algorithmic}	
	\caption{Bottom-Up Compositional Clustering}\label{alg:hmp}	
\end{algorithm}

%% file: greedy.tex
    \section{Greedy structure learning}
    \label{sec:gcc}
    In this section, we describe a greedy structure learning algorithm that infers the full dependency structure of the multi-layer CABM from natural images. Figure~\ref{fig:cabm_learn} illustrates the two phases of this learning process: A bottom-up compositional clustering process (Figure~\ref{fig:cabm_learn} blue box) and a top-down model composition phase (Figure~\ref{fig:cabm_learn} green box). In the bottom-up process, the parts of lower layers in the hierarchy are learned first and subsequently composed into higher-order parts. The top-down process composes the (independent) hierarchical parts into a holistic object model. The following paragraphs describe both processes in detail.

    \begin{figure*}
        \centering
        \subfloat[\label{fig:evo-learning}]{\includegraphics[width=0.9\textwidth]{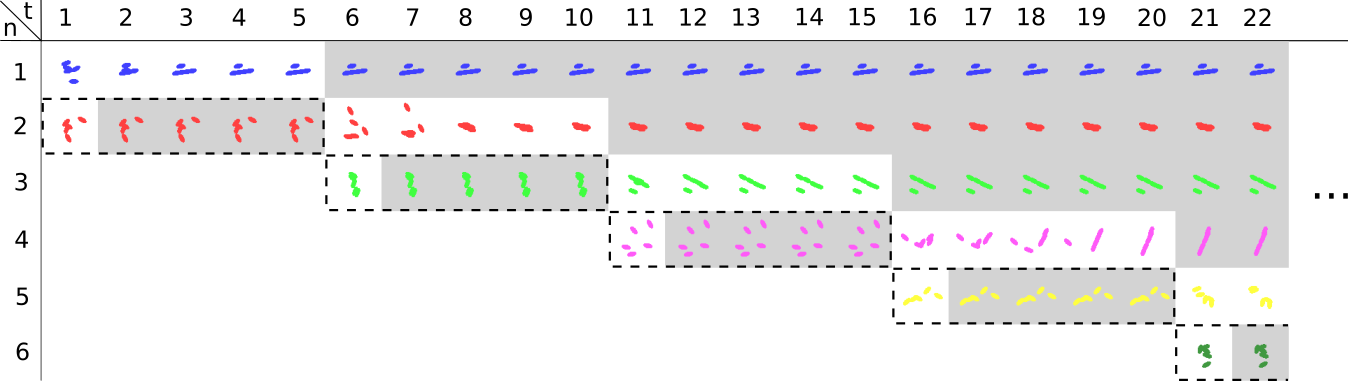}}\quad
        \subfloat[\label{fig:evo-encoding}]{\includegraphics[height=3cm]{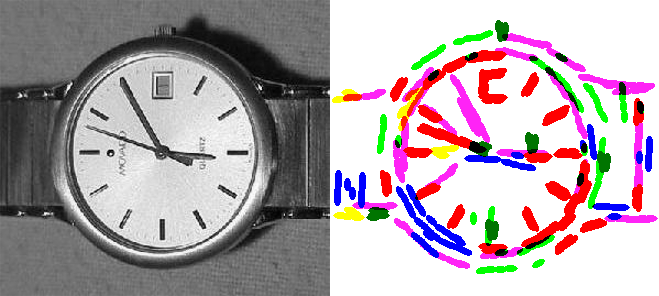}}\quad
	    \caption{Illustration of the proposed greedy EM-type learning process. The part models are composed of $5$ Gabor filters which are represented as colored ellipses. (a) The first $t=22$ iterations of the greedy learning scheme. Each row shows the evolution of a part model over time. Each column shows the learning result at one iteration of the learning process. When a new part is initialized ($t=1,6,11,\dots$), also a generic background model is learned from the training image (marked by dashed rectangles). The background model and the learned part models are not adapted in the subsequent iterations (gray background) but serve as competitors for data in the E-step. For more details refer to Section \ref{sec:bu-learning}. (b) An example encoding of a training image with the learned part models.}
	\label{fig:evo}
    \end{figure*}       	
  	\subsection{Bottom-up compositional clustering}
  	\label{sec:bu-learning}
  	We propose to formulate the structure learning task as a compositional clustering process that is described in Algorithm \ref{alg:hmp}. The dependence structure of our HCM is a tree-like Markov random field (Figure \ref{fig:cabm_graph} \& Equation \ref{eq:mcabm}). This enables us to learn the model in a bottom-up manner, i.e. we can learn the first-layer parts first, before proceeding to compose them into larger parts (Algorithm \ref{alg:hmp}, lines $1-5$). At each layer of the hierarchy, the parts are learned according to a \textit{greedy EM-type learning scheme} that infers the structure of each part as well as the number of part models from training images $I$ (Algorithm \ref{alg:hmp}, line $3$).
 	 
 	\textbf{Greedy EM-type learning.} The general procedure of the proposed greedy learning scheme is described in (Algorithm \ref{alg:hmp}, line $6-19$), while Figure \ref{fig:evo-learning} analogously illustrates a particular learning example. We first explain the algorithmic process and then comment on the visual illustration. The learning process is initialized with two part models ($\beta^1_1, \beta^1_2$) which are learned with matching pursuit \cite{mallat1993} from image patches that are randomly sampled from the training data (Algorithm \ref{alg:hmp}, line $10$). Subsequently an EM-type update scheme is performed (Algorithm \ref{alg:hmp}, line $12-16$) as follows:
 	\begin{enumerate}
 		\item  \textbf{Detection (E-step)}: Detect part models in the training images at different locations and orientations. Cut out patches at the detected positions which serve as new training data for the M-step (Algorithm \ref{alg:hmp}, line $14$).  		
 		\item  \textbf{Learning (M-step)}: Learn a part model from the training patches with matching pursuit \cite{mallat1993} (Algorithm \ref{alg:hmp}, line $16$).
 	\end{enumerate}
 	
	During the EM iterations we only update one part model ($\beta^1_1$) while the other model ($\beta^1_2$) stays fixed to its initial state and only participates in the detection phase. In doing so, it serves as a generic background model preventing $\beta^1_1$ from explaining image patches for which the normalized posterior $p(\beta^1_1|I)$ is smaller than $p(\beta^1_2|I)$ (analogous to Equation \ref{eq:mcabm} and Figure \ref{fig:cabm_graph-a}). This mechanism supports $\beta^1_1$ in specializing to a particular image structure (see e.g. the ticks of the watch in Figure \ref{fig:cabm_learn}) by explaining away irrelevant data (e.g. patches sampled from background clutter).	After a fixed number of iterations and two new part models $\beta^1_2$ and $\beta^1_3$ are added to the model pool (Algorithm \ref{alg:hmp}, line $10$). This time, however, the training patches are not sampled randomly, but inversely proportional to the marginal posterior $p(\beta^1_1|I)$. In this way, those regions which are well explained by the already learned model $\beta^1_1$, are less likely to be sampled as training data for the new models. In the following iterations, $\beta^1_2$ is updated in the learning phase, while $\beta^1_1$ and $\beta^1_3$ serve as competitors, explaining away irrelevant training patches in the detection phase. The learning proceeds until iteration $\beta^1_2$ is converged. This greedy learning scheme repeats until a newly initialized model is not able to explain training patches better than any previously learned model or the background model (Algorithm \ref{alg:hmp}, line $19$). 
	
	Figure \ref{fig:evo-learning} analogously illustrates the first iterations of such a greedy EM-type learning scheme. In the illustrated example, first-layer parts $\beta^1_i$ are learned to be composed of five Gabor filters (colored ellipses). We can observe that the learned models specialize to a particular local image structure, while the background models (dashed rectangles) have a rather random structure. Figure \ref{fig:evo-encoding} illustrated an encoding of a training image with the final set of part models $B^1=\{\beta^1_n | n=1,\dots,N_1\}$. Note how different models have specialized to different parts of the watch.

    \textbf{Bottom-up learning.} After the parts of the first layer $B^1$are learned, the structure induction process continues by composing the elements of $B^1$ into parts of the second layer $B^2$. Thereby, we follow the same greedy EM-type learning process. This time, however, instead of composing single basis filter, the algorithm composes the elements of $B^1$ into higher order parts. We repeat the compositional learning iteratively layer by layer until the normalized model posterior does not increase anymore (Algorithm \ref{alg:hmp}, line $5$), thus generating dictionaries of hierarchical part models at each layer of the hierarchy $\{B^1,\dots,B^L\}$ (Figure \ref{fig:cabm_learn}, blue box).

	\subsection{Top-down model building} 

	After the bottom-up learning process, the learned part dictionaries $\{B^1,\dots,B^L\}$ must be composed into a holistic object model (Figure \ref{fig:cabm_learn}, green box). Note that different parts of an object can terminate at different layers of the hierarchy. For example, the hour markings on the dial of the watch in Figure \ref{fig:cabm_learn} (orange box) are represented at the second layer, whereas, the circular shape of the watch is composed of more elements and is therefore represented at a higher layer. We suggest a top-down model building process for learn the dependency structure of the complete object, we introduce a top-down model building process. 
	
	The training images are first aligned, by detecting the part model of the highest layer $B^L$ in all training images followed by aligning the images such that the models $B^L$ are in a canonical orientation and position. After this alignment step, we proceed in a top-down manner (green box in Figure \ref{fig:cabm_learn}), adding parts from the highest layer to the object model with matching pursuit. We iteratively proceed layer-wise until the bottom layer of the hierarchy. At this point, we have learned a hierarchical compositional object model from natural images (orange box in Figure \ref{fig:cabm_learn}). Note that the number of layers $L$ , the number of parts per layer $N_L,\dots,N_0$ and the hierarchical dependency structure have been learned from natural images without restricting the object's geometry or requiring segmented training data.

   	\input{exp_tables}

%% file: exp_tables.tex
	\begin{table*}
			\centering
			\small
			\scalebox{0.9}{
				\begin{tabular}{|c|c|c|c|c|c|c|c|c|}
					\hline
					Methods & C $\rightarrow$ A & C $\rightarrow$ D & A $\rightarrow$ C & A $\rightarrow$ W & W $\rightarrow$ C & W $\rightarrow$ A & D $\rightarrow$ A & D $\rightarrow$ W \\
					\hline
					KSVD \cite{aharon2006} & 20.5 $\pm$ 0.8 & 19.8 $\pm$ 1.0 & 20.2 $\pm$ 0.9 & 16.9 $\pm$ 1.0 & 13.2 $\pm$ 0.6 & 14.2 $\pm$ 0.7 & 14.3 $\pm$ 0.3 & 46.8 $\pm$ 0.8 \\
					SGF \cite{gopalan2011domain} & 36.8 $\pm$ 0.5 & 32.6 $\pm$ 0.7 & 35.3 $\pm$ 0.5 & 31.0 $\pm$ 0.7 & 21.7 $\pm$ 0.4 & 27.5 $\pm$ 0.5 & 32.0 $\pm$ 0.4 & 66.0 $\pm$ 0.5 \\
					HABM \cite{dai2014} & 53.7 $\pm$ 4.7 & 43.2 $\pm$ 4.9  & 41.2 $\pm$ 1.6  & 28.1 $\pm$ 2.0  & 25.8 $\pm$ 1.6  & 33.5 $\pm$ 2.9  & \textbf{34.6 $\pm$ 3.7} & \textbf{68.2 $\pm$ 2.9}\\
					OURS & \textbf{62.3 $\pm$ 3.4} & \textbf{43.7 $\pm$ 2.9} & \textbf{54.0 $\pm$ 2.4} & \textbf{33.3 $\pm$ 1.7} & \textbf{29.5 $\pm$ 1.1} & \textbf{35.0 $\pm$ 3.6} &  33.1 $\pm$ 2.4 & 65.6 $\pm$ 3.8 \\
					\hline
				\end{tabular}
			}
			\caption{Unsupervised domain adaptation: Classification scores on the Four Domain Dataset. The four domains are Amazon (A), Webcam (W), Caltech256(C), DSLR (D).  We compare our results to dictionary learning with K-SVD, subspace geodesic flow (SGF), and the hierarchical active basis model (HABM). Our approach outperforms other generative approaches in six out of eight experiments.}
			\label{table:unsupervised}
    \end{table*}
    
	\begin{table*}	
			\centering
			\small	
		\scalebox{0.89}{
				\begin{tabular}{|c|c|c|c|c|c|c|c|c|}
					\hline
					Methods & C $\rightarrow$ A & C $\rightarrow$ D & A $\rightarrow$ C & A $\rightarrow$ W & W $\rightarrow$ C & W $\rightarrow$ A & D $\rightarrow$ A & D $\rightarrow$ W \\
					\hline
					Metric \cite{saenko2010} & 33.7 $\pm$ 0.8 & 35.0 $\pm$ 1.1 & 27.3 $\pm$ 0.7 & 36.0 $\pm$ 1.0 & 21.7 $\pm$ 0.5 & 32.3 $\pm$ 0.8 & 30.3 $\pm$ 0.8 & 55.6 $\pm$ 0.7 \\
					SGF \cite{gopalan2011domain} & 40.2 $\pm$ 0.7 & 36.6 $\pm$ 0.8 & 37.7 $\pm$ 0.5 & 37.9 $\pm$ 0.7 & 29.2 $\pm$ 0.7 & 38.2 $\pm$ 0.6 & 39.2 $\pm$ 0.7 & 69.5 $\pm$ 0.9 \\
					FDDL \cite{yang2011fisher} & 39.3 $\pm$ 2.9 & 55.0 $\pm$  2.8 & 24.3 $\pm$ 2.2 & 50.4 $\pm$ 3.5 & 22.9 $\pm$ 2.6 & 41.1 $\pm$ 2.6 & 36.7 $\pm$ 2.5 & 65.9 $\pm$ 4.9 \\
					HMP \cite{bo2011} & 67.7 $\pm$ 2.3 & 70.2 $\pm$  5.1 & 51.7 $\pm$ 4.3 & 70.0 $\pm$ 4.2 & 46.8 $\pm$ 2.1 & 61.5 $\pm$ 3.8 & 64.7 $\pm$ 2 & 76.0 $\pm$ 4 \\
					SDDL \cite{shekhar2013} & 49.5 $\pm$ 2.6 & \textbf{76.7 $\pm$ 3.9} & 27.4 $\pm$ 2.4 & \textbf{72.0 $\pm$ 4.8} & 29.7 $\pm$ 1.9 & 49.4 $\pm$ 2.1 & 48.9 $\pm$ 3.8 & 72.6 $\pm$ 2.1 \\
					HABM \cite{dai2014} & 68.3 $\pm$ 2.3 & 57.4 $\pm$ 6.0 & 52.7 $\pm$ 3.0 & { 54.8 $\pm$ 2.8} & 42.2 $\pm$ 3.1 & 57.1 $\pm$ 3.5 & 60.1 $\pm$ 3.2 & \textbf{79.7 $\pm$ 2.5} \\
					OURS & \textbf{72.2 $\pm$ 0.7}& 58.1 $\pm$ 5.1 & \textbf{58.5 $\pm$ 1.2} & 53.4 $\pm$ 1.2 & \textbf{47.6 $\pm$ 1.8} & \textbf{61.7 $\pm$ 3.2} &\textbf{ 65.6 $\pm$ 2.8} & 78.5 $\pm$ 2.0 \\
					\hline
				\end{tabular}
			}
		\caption{Semi-supervised domain adaptation: Classification scores on the Four Domain Dataset. The four domains are Amazon (A), Webcam (W), Caltech256(C), DSLR (D). We compare our results to subspace geodesic flow (SGF), fisher discriminant dictionary learning (FDDL), shared	domain-adapted dictionary learning, hierarchical matching pursuit (HMP), and the hierarchical active basis model (HABM). Our approach outperforms the other approaches in five out of eight experiments.}
		\label{table:supervised}
	\end{table*}   

%% file: results.tex
	\vspace{-0.1cm}
	\section{Results}
	\label{sec:exp}
	We evaluate the proposed HCM learning scheme qualitatively by comparing it to the HABM approach proposed in \cite{dai2014}. Quantitative results are presented at the task of domain adaptation on the Four Domain Dataset \cite{gong2012} and compared to other generative approaches. Note that it is difficult to evaluate generative object models at object recognition tasks, as they are optimized via a data reconstruction criterion and thus naturally perform worse than methods which are directly optimized via a discriminative criterion. Furthermore, generative models provide a manifold of information in addition to the mere class label, such as e.g. the position of the object, detailed part annotations and a foreground-background segmentation.
	
	\textbf{Parameter settings.} In our experiments, the images have a mean height of $300$ pixels while a Gabor filter has a quadratic size of $17$ pixels. The Gabor filters and higher layer parts are rotated in 10 degree steps. 
	We found empirically that in the greedy learning scheme, a part model is converged to stable solution after $5$ learning iterations. The hierarchical graph structure is defined to compose two parts at each layer of the hierarchy. Changing the number of parts to be composed at each layer would have implications on the overall number of layers learned during training, however, we found that it has no particular impact on the overall performance in the quantitative experiments.
	
	    \begin{figure*}
	    	\centering
	    	\subfloat[\label{fig:unsupinp}]{\includegraphics[height=5.4cm]{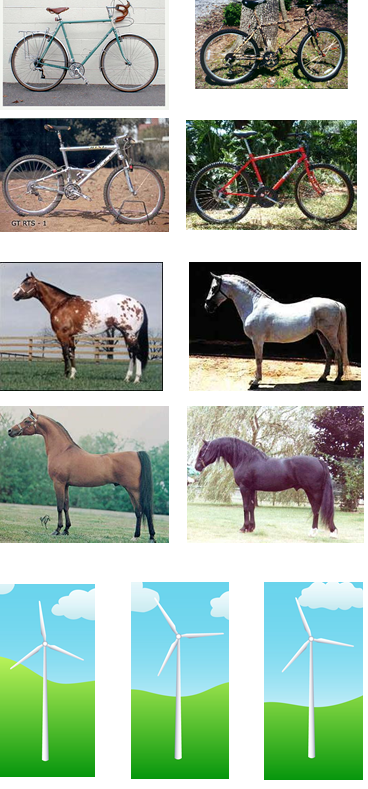}}\quad
	    	\subfloat[\label{fig:unsupbot}]{\includegraphics[height=5.4cm]{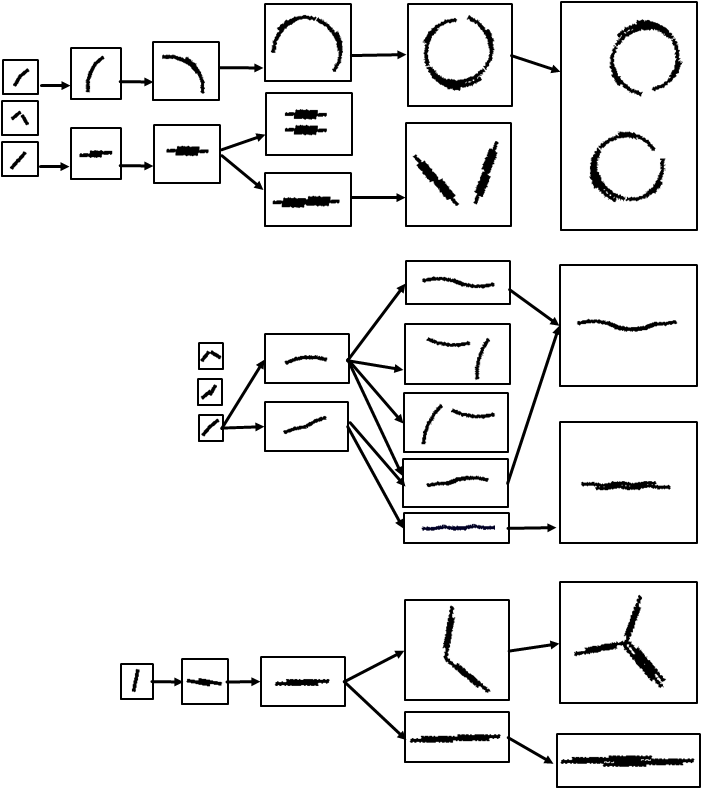}}\quad
	    	\subfloat[\label{fig:unsuptop}]{\includegraphics[height=5.4cm]{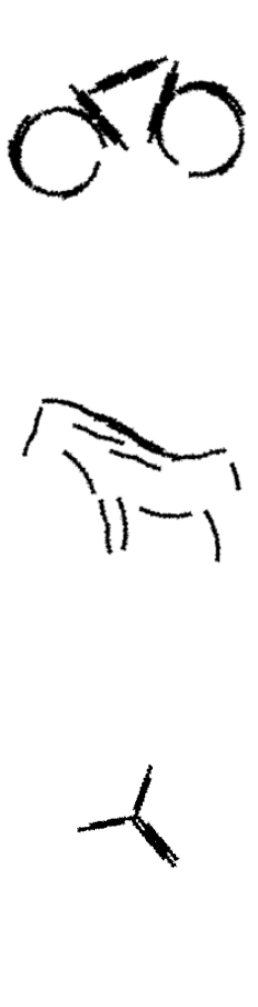}}\quad
	    	\subfloat[\label{fig:habm}]{\includegraphics[height=5.4cm]{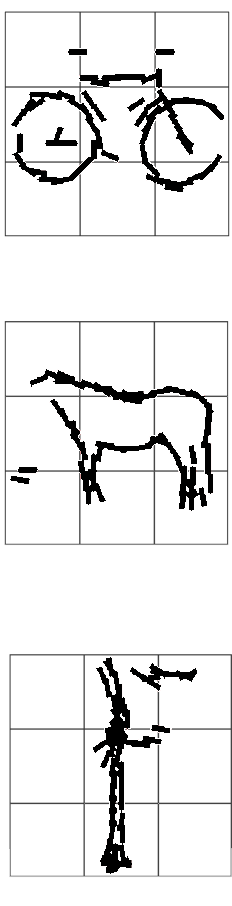}}\quad		        
	    	\caption{Learned hierarchical compositional models. (a) Samples from the training data. (b) The hierarchical part dictionary learned with our the bottom-up process. (c) The holistic object model after the top-down process. (d) The HCM learned with the HABM approach \cite{dai2014}. The gray squares indicate the parts of their HCM. Compared to the HABM, our method is able to learn the number of parts and layers of the hierarchy. Both approaches are not able to learn the holistic structure of the windmill due to the strong relative rotation between its parts.}
	    	\label{fig:unsupResult}
	    \end{figure*}	
	\subsection{Visual Domain Adaptation}
	We follow the common evaluation protocol of using generative part-based models as feature extractors for discriminative methods. The rationale behind this setup is that generative models are highly adaptive to changes in the visual environment and thus are suited as feature extractors for domain adaptation tasks. We evaluate our approach on the Four Domain Dataset \cite{gong2012}, which is composed of $10$ identical classes from the following datasets: \emph{Amazon} with images downloaded from Amazon; \emph{DSLR} with high-resolution images; \emph{Webcam} with low-resolution images and images from \emph{Caltech256} \cite{griffin2007}. In each dataset, the image resolution, lighting conditions, background, the object textures and positions in space vary significantly. We follow the standard evaluation protocol as introduced in \cite{gong2012}. We test two experimental setups: In the semi-supervised setting, the algorithm has access to a small amount of data from the target domain, whereas in the unsupervised setting the training images are only sampled from the source domain. As proposed in \cite{dai2014}, we use the learned HCM as feature extractor for a spatial pyramid matching \cite{lazebnik2006}. A multi-class SVM is trained on the extracted features and used for classification. The classification results in Tables \ref{table:unsupervised} \& \ref{table:supervised} show that:
	
	\textbf{Our approach outperforms other generative methods at the task of unsupervised domain adaptation} (Table \ref{table:unsupervised}). Note that our method uses exactly the same Gabor basis as the HABM. The performance increase can be attributed to the fact that we learn the hierarchical structure and do not specify it a-priori as in the HABM. Interestingly, our approach is outperformed by HABM when learning from the DSLR dataset, which has a strong intra-domain viewpoint variation. As we enforce the learning of a single holistic object model, our approach specializes to one of the viewpoints, whereas, in the HABM method multiple templates are learned. Note that our approach could in principle be extended to multi-object learning.

	\textbf{Our models achieve the best performance in most experiments in the semi-supervised setting}  (Table \ref{table:supervised}). Again we outperform the HABM approach in most experiments. Interestingly, the performance gap between our method and HABM when learning from DSLR is essentially closed in the semi-supervised setting. The reason is that the objects in the Amazon and Webcam class have significantly less variation in the viewpoint. Our learning scheme can leverage this and will specialize to the most common viewpoint in the data. Therefore, the advantage of having multiple templates in the HABM is reduced.
	
	\subsection{Qualitative Evaluation}
	\textbf{Our approach learns the structure of HCMs from natural images.}	The learning results of Figure \ref{fig:unsupbot} \& \ref{fig:unsuptop} demonstrate that our approach is able to learn the hierarchical structure of HCMs from cluttered natural images. Importantly, our approach does not depend on detailed human supervision during learning. This is in contrast to prior work that relies on detailed object segmentations \cite{george2017generative,kortylewski2016probabilistic} or a-priori knowledge about the hierarchical structure~\cite{dai2014,zhu2008}.
	
	\textbf{Our HCMs are more efficient and semantically more meaningful compared to prior work.} Learning the full hierarchical structure enables the reuse of parts within the hierarchy (e.g. the wheels of the bike in Figure \ref{fig:unsuptop}), which is not possible in the HABM approach \cite{dai2014} (Figure \ref{fig:habm}). Therefore, our HCMs have semantically more meaningful parts that provide additional information about the internal semantics of the object. Furthermore, our learning process is more data efficient as the part models can leverage the redundancy within objects (e.g. if the same part occurs multiple times within the same object).
	
	A limitation of our approach and any prior work, including HABM, is that so far it is not possible to learn HCMs of articulated objects (e.g. the windmill in Figure~\ref{fig:unsupResult}). Although the individual parts of the windmill are learned by our bottom-up process (Figure \ref{fig:unsupbot}), the top-down process cannot compose the parts into a holistic object model as our deformation model assumes that the relative orientation between parts of an object stays approximately the same. 

%% file: conclusion.tex
	\section{Conclusion}
	\vspace{-.17cm}
	 In this work, we considered the challenging problem of learning a hierarchical generative model of an object from only a set of images which show examples of the object in the presence of variable background clutter. In this context, we made the following contributions:
	 
	\textbf{Multi-layer Compositional Active Basis Models (CABMs).} Building on related work, we proposed a generalized probabilistic formulation of CABMs with arbitrary numbers of layers and parts per layer. Our model is more flexible and enables the representation of objects with very different geometry structures. It also opens the possibility to learn hierarchical object representations which efficiently re-use parts and thus provide rich information about the objects internal structure (Fig. \ref{fig:2ours} \& \ref{fig:unsupResult}).
	
	\textbf{Structure learning from cluttered data.} We introduced a framework for learning the structure of multi-layer CABMs from natural images based on class labels only. Notably, we were able to learn the full dependency structure, including the number of layers in the hierarchy and the number of parts per layer, despite complex variations in the images, in terms of highly variable background clutter and object appearance. Importantly, our framework overcomes the limitations of related works which either require segmented training data or make too strong assumptions about the object's geometry. The learned models also outperformed other generative object models at object classification on a standard domain transfer dataset. 
	
	\footnotesize{\textbf{Acknowledgements:} A.K. is supported by the Novartis University of Basel Excellence Scholarship for Life Sciences and the SNSF grant P2BSP2\_181713. A.W., M.W., S.P. are partially supported by the NCCR MARVEL and SNSF grants CR32I2159682 and 51MRP0158328.}